\pdfoutput=1

\documentclass[11pt]{article}
\usepackage[table]{xcolor}
\usepackage[]{acl}

\usepackage{times}
\usepackage{latexsym}
\usepackage{enumitem}

\usepackage[T1]{fontenc}

\usepackage[utf8]{inputenc}

\usepackage{microtype}

\usepackage{inconsolata}
\usepackage{threeparttable}

\usepackage{graphicx}

\usepackage{changepage}   %
\usepackage{booktabs}
\usepackage{subcaption}

\usepackage{listings,multicol}

\usepackage{diagbox}

\usepackage{float}

\usepackage[T1]{fontenc}

\colorlet{punct}{red!60!black}
\definecolor{background}{HTML}{EEEEEE}
\definecolor{delim}{RGB}{20,105,176}
\colorlet{numb}{magenta!60!black}

\lstdefinelanguage{json}{
    showspaces=false,
    showtabs=false,
    breaklines=true,
    postbreak=\raisebox{0ex}[0ex][0ex]{\ensuremath{\color{gray}\hookrightarrow\space}},
    breakatwhitespace=true,
    basicstyle=\ttfamily\small,
    upquote=true,
    frame=lines,
    morestring=[b]",
    backgroundcolor=\color{background},
    literate=
     *{0}{{{\color{numb}0}}}{1}
      {1}{{{\color{numb}1}}}{1}
      {2}{{{\color{numb}2}}}{1}
      {3}{{{\color{numb}3}}}{1}
      {4}{{{\color{numb}4}}}{1}
      {5}{{{\color{numb}5}}}{1}
      {6}{{{\color{numb}6}}}{1}
      {7}{{{\color{numb}7}}}{1}
      {8}{{{\color{numb}8}}}{1}
      {9}{{{\color{numb}9}}}{1}
      {\{}{{{\color{delim}{\{}}}}{1}
      {\}}{{{\color{delim}{\}}}}}{1}
      {[}{{{\color{delim}{[}}}}{1}
      {]}{{{\color{delim}{]}}}}{1},
}

\lstdefinestyle{pseudocode}{
    basicstyle=\ttfamily,
    keywordstyle=\color{blue}\bfseries,
    identifierstyle=\color{black},
    commentstyle=\color{green!50!black},
    stringstyle=\color{red},
    numbers=left,
    numberstyle=\tiny\color{gray},
    numbersep=5pt,
    breaklines=true,
    showstringspaces=false,
    frame=single,
    captionpos=b,
    language={}
}

\title{Causal Micro-Narratives}

\author{
Mourad Heddaya \\ University of Chicago \\  \scalebox{0.87}[0.9]{{\tt {mourad@uchicago.edu}}} \And 
Qingcheng Zeng \\ Northwestern University \\  \scalebox{0.87}[0.9]{{\tt {qingchengzeng2027@u.northwestern.edu }}} \And
Chenhao Tan \\ University of Chicago \\  \scalebox{0.87}[0.9]{{\tt {chenhao@uchicago.edu}}}
\AND
Rob Voigt \\ Northwestern University \\  \scalebox{0.87}[0.9]{{\tt{robvoigt@northwestern.edu}}} \And
Alexander Zentefis \\ Hoover Institution, Stanford University\\  \scalebox{0.87}[0.9]{{\tt{zentefis@stanford.edu}}}
}

\begin{document}
\maketitle
\begin{abstract}

We present a novel approach to classify \emph{causal micro-narratives} from text. These narratives are sentence-level explanations of the cause(s) and/or effect(s) of a target subject. The approach requires only a subject-specific ontology of causes and effects, and we demonstrate it with an application to inflation narratives. Using a human-annotated dataset spanning historical and contemporary US news articles for training, we evaluate several large language models (LLMs) on this multi-label classification task. The best-performing model---a fine-tuned Llama 3.1 8B---achieves F1 scores of 0.87 on narrative detection and 0.71 on narrative classification. Comprehensive error analysis reveals challenges arising from linguistic ambiguity and highlights how model errors often mirror human annotator disagreements. This research establishes a framework for extracting causal micro-narratives from real-world data, with wide-ranging applications to social science research.\footnote{Data is available at \url{https://mheddaya.com/research/narratives}}

\end{abstract}

\section{Introduction}

In recent years, social scientists have increasingly recognized the power of narratives (i.e., popular stories about economic, political, or social topics) to shape individual and collective behavior. These narratives can influence people's beliefs and decisions---like when to invest in the stock market, buy a home, or pursue higher education---and can quickly spread through the collective consciousness. Nobel Prize-winning economist Robert Shiller argues that if we fail to consider and understand the properties of narratives, ``we remain blind to a very real, very palpable, very important mechanism for economic change, as well as a crucial element for economic forecasting'' (\citealp{shiller2017narrative}).

\begin{figure}[t!]
    \centering
    \includegraphics[width=0.48\textwidth]{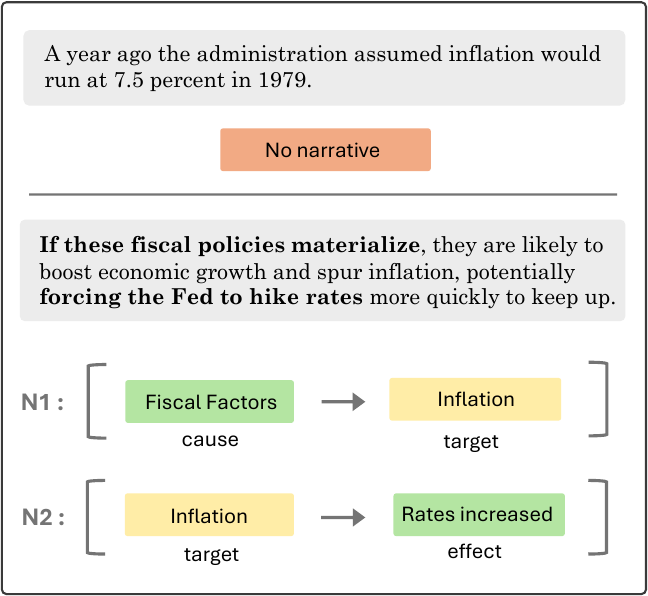}
    \caption{Causal micro-narrative classification task examples for the \textit{target} `inflation.' In the first sentence, no narratives are identified; in the second, two narratives (N1 and N2) are identified, one representing a cause of the \textit{target} and the other representing an effect of it.
}
    \label{fig:mainfig}
\end{figure}

While the importance of narratives has become well recognized, formulating an operational definition remains challenging. Recent work in economics and psychology has proposed definitions based on how narratives affect people's sentiment or moral reasoning~\citep{flynn2022macroeconomics,benabou2018}, while other research in these fields has proposed definitions based on a \emph{causal} account of events \citep{akerlof2016bread,eliaz2020model,kendall2022causal,morag2023narratives,andre2023narratives,barron2023narrative}. These works capture important aspects of narratives, but they do not propose methods to uncover narratives from real-world data. Because narratives are disseminated to broad audiences through free-form formats like text and speech (e.g., printed, television, or web media), it is challenging to systematically extract them and quantify their prevalence and influence. 

This paper aims to address both these conceptual and technical challenges. We introduce the concept of \textit{causal micro-narratives}, along with a multi-label classification task to extract them from text. We define \textit{causal micro-narratives} as sentence-level explanations of the cause(s) and/or effect(s) of a target subject (e.g., an event, occurrence, emotion, phenomenon). These micro-narratives are pervasive in everyday communication. When people speak and write, they often explicitly or implicitly propose causal relations between entities and outcomes that reflect their understanding of how the world works. For instance, if someone were to say, ``Jane is tired, so she won't make it to the show tonight,'' they implicitly propose a ``micro'' story that frames Jane's tiredness as the cause and her absence as the effect.

As an application of this concept, we choose \emph{inflation} as the target that centers the micro-narratives we examine. Inflation is a popular and salient topic in news media, and can be clearly summarized by a single word, which aids in data filtering. Figure \ref{fig:mainfig} illustrates how our framework distinguishes a sentence conveying a micro-narrative about inflation and one that does not. The top sentence simply reports factual news about inflation, whereas the bottom one presents two causal claims: (1) ``fiscal policies" will cause inflation, and (2) the Federal Reserve will increase interest rates in response to (i.e., as an effect of) inflation. We label these two micro-narratives  \textit{fiscal factors} and \textit{rates increased}, respectively. 

We propose an ontology of causes and effects of inflation, and we create a large scale dataset of causal micro-narratives according to this ontology, classifying sentences from contemporary and historical U.S. news articles. We start with a subset of human annotations, and then use them to train various models for classifying these narratives at scale. The best model achieves F1 scores as high as 0.71, despite the difficulty of the task, having 18 classes that in some cases are semantically similar. Our comparison of different models reveal that smaller fine-tuned large language models (LLMs) outperform larger models like GPT-4o, while also being more scalable and cost efficient.

To better characterize our dataset and the performance of our classifiers, we conduct an in-depth error-analysis of inter-annotator disagreements and the in- and out-of-domain generalization of each evaluated model. Furthermore, we identify and cross-reference systematic classification errors with annotator disagreements. We find that the best-performing fine-tuned LLMs have a small performance degradation on out-of-domain data, but overall are robust to domain shifts across texts that are written 50 years apart. The errors produced by LLMs that are fine-tuned on our human-annotated data reflect the natural disagreements between annotators to a far greater extent than the errors produced by GPT-4o in a few-shot, in-context learning setting.

In summary, we make the following contributions:

\begin{enumerate}[leftmargin=*, itemsep=-2pt, topsep=0pt]
    \item We introduce and define the concept of causal micro-narratives, presenting a novel task for extracting them from real-world text.
    
    \item We curate a dataset of annotated inflation-related causal micro-narratives from both historical and contemporary U.S. news articles.
    
    \item We develop and demonstrate methods for effectively automating narrative classification at scale, making publicly available fine-tuned LLMs for this purpose. Additionally, we showcase robust out-of-domain performance of these models.
    
    \item We conduct a comprehensive error analysis, revealing systematic similarities between model classifications and human annotation disagreements. This analysis highlights the task's complexity and identifies potential inherent ambiguities.
\end{enumerate}

\section{Related Work}

\subsection{Definitions and Theoretical Frameworks}

Early work by \citet{labov1997narrative} defined narratives as temporal accounts of event sequences, providing a formal framework for analyzing personal narratives. Building on this, \citet{akerlof2016bread} expanded the definition to include causally linked events and their underlying sources, emphasizing the role of narratives in decision-making processes.

More recent work has further refined these concepts. \citet{eliaz2020model} represent narratives as directed acyclic graphs (DAGs), drawing on Bayesian Networks to model the equilibrium of narratives. \citet{shiller2017narrative} likened narratives to viral phenomena, defining them as interpretive stories about economic events that spread contagiously. \citet{benabou2018} focused on the persuasive aspect of narratives in moral decision-making, while \citet{flynn2022macroeconomics} emphasized their contagious nature in belief formation.

\citet{morag2023narratives} and \citet{barron2023narrative} both highlight the causal and interpretive aspects of narratives. The former defines narratives as stories that establish causal links between events on a timeline, while the latter views them as subjective explanations of datasets, particularly in the context of persuasion.

\subsection{Methodological and Empirical Studies}

Studies have proposed different methodologies to empirically measure economic narratives. \citet{JALIL201626} analyze word frequency in newspapers and forecasts to study inflation expectations during the Great Depression. More advanced NLP techniques have been applied as well. \citet{lange2022towards} extended the RELATIO method of \citet{ash2021relatio} to extract narratives based on \citet{Roos_Reccius_2021}'s definition. \citet{gueta2024llmslearnmacroeconomicnarratives} try to leverage LLMs to extract and summarize economic narrative from tweets. However, they do not clearly define \textit{economic narrative} nor do they evaluate the LLM's performance. \citet{flynn2022macroeconomics} utilize sentiment analysis on firm 10-K filings to build a macro model explaining economic fluctuations.

\citet{andre2023narratives} use open-ended surveys and DAGs to study narratives around recent high U.S. inflationary period. They contrast the narratives that households and experts write down, finding that household narratives significantly shape expectations. Their work also include experiments manipulating narratives to measure their impact on inflation expectations.

\citet{ali2021causality} survey the broader field of causality extraction from text. Most causality extraction tasks are general domain, but existing methods are not very robust to complex sentence structures. Recent work by \citet{sun-etal-2024-event} proposes a promising prompt-based technique with large language models to extract causal relationships in fictional stories instead of news text.

\section{Causal Micro-Narratives}

We define a \textit{causal micro-narrative} as
\vspace{5pt}
\begin{adjustwidth}{0.75cm}{0.75cm}
\textit{a sentence-level explanation of the cause(s) and/or effect(s) of a target subject}.
\end{adjustwidth}
\vspace{5pt}

The term ``narrative'' is most commonly applied to the discourse-level conception of story-telling that depicts  sequences of events, usually in long-form texts \cite[e.g.,][]{piper-2023-computational}. By contrast, here we focus on narrative fragments within individual sentences, which can capture stories about implicit and explicit cause-effect relationships that people express as they speak or write, sometimes in subtle or subconscious ways.
Recent work in cognitive science highlights the prevalence of causal connectives in English and how they reveal the importance of causal relationships in the way we think and express ourselves \citep{iliev2016morecausality, brown1983237, SandersSweetser2009}.

\subsection{Narrative Classification Task}
\label{sec:task}

We propose a narrative classification task that operationalizes our definition of \textit{causal micro-narratives}. Unlike the more general task of causality mining \cite{ali2021causality}, we suggest that a productive approach to capturing how such micro-narratives accumulate at scale should be domain-specific. Specifically, we propose a framework in which we first identify a \textit{target} about which we hope to capture micro-narratives. Conceptually a target can by any entity, event, or phenomenon of interest.

Then, we define an ontology of the causes that can lead to that target and the effects that can follow from it.  Thus, the narrative classification task is to identify, according to the ontology, sentences that express a narrative about the target subject and to predict the particular cause(s) and/or effect(s) related to the target that are present.

\subsection{Case Study: Inflation Narratives}

As an application of this definition and for the purposes of this paper, we focus specifically on \textit{inflation} as the target. We develop an ontology, presented in Table \ref{tab:narratives-labels}, consisting of 8 causes of inflation and 11 effects that could follow from inflation. The causes and effects were curated by an expert economist based on domain knowledge and researching relevant resources online. See Appendix \ref{sec:cls-task} for additional details on this process, and detailed descriptions of all the causes and effects. Ultimately, we setup the following classification task: given a sentence, identify (1) whether the sentence expresses a narrative about inflation, and (2) the expressed cause(s) and/or effect(s) of the inflation.

For this case study, we choose a target event that is fairly unambiguously summarized by a single word, \textit{inflation}, which allows for straightforward data filtering. Nonetheless, the causal micro-narrative classification task could be applied to target events or phenomena that are expressed in more varied ways, but this would introduce more complicated filtering strategies or an additional preliminary event extraction step.

\begin{table*}[htbp]
\centering

\begin{tabular}{p{8cm}p{7.5cm}}
\toprule
\textbf{Causes (label)} & \textbf{Effects (label)} \\
\midrule
Demand-side Factors (demand) & Reduced Purchasing Power (purchase) \\
Supply-side Factors (supply) & Cost of Living Increases (cost) \\
Built-in Wage Inflation (wage) & Uncertainty Increases (uncertain) \\
Monetary Factors (monetary) & Interest Rates Raises (rates) \\
Fiscal Factors (fiscal) & Income or Wealth Redistribution (redistribution) \\
Expectations (expect) & Impact on Savings (savings) \\
International Trade \& Exchange Rates (international) & Impact on Global Trade (trade) \\
Other Causes (other-cause) & Cost-Push on Businesses (cost-push) \\
& Social and Political Impact (social) \\
& Government Policy \& Public Finances Impact (govt) \\
& Other Effects (other-effect) \\
\bottomrule
\end{tabular}

\caption{Inflation Narrative Causes and Effects. The \textbf{label} in parentheses refers to the abbreviated name used during classification in both few-shot and fine-tuning experiments. See Appendix \ref{tab:narratives} for additional details.}
\label{tab:narratives-labels}
\end{table*}

\section{Dataset}

We use two data sources in our investigation of inflation narratives in news: NOW Corpus for contemporary news data \citep{davies2016now} and ProQuest for historical data. We selected these datasets because their differences allow us to assess the generalizability of our task and the classification methods we test. The articles in each dataset were written roughly 50 years apart and the NOW corpus includes a high degree of stylistic variation, as the articles are sourced from a range of online sources.

For each dataset, we segment articles into sentences and filter sentences that contain the keyword ``inflation". Filtering allows us to focus on relevant sentences,
 enabling us to efficiently target our human annotations, as well as reduce the total number of sentences to a more computationally feasible quantity.

\subsection{Contemporary News: NOW Corpus}

We use data from the NOW Corpus covering 2012-2023. The dataset consists of online news articles, which we filter to only include U.S. articles written in English. The final filtered dataset, including ``inflation'' keyword filtering, contains 118,383 articles and 284,220 sentences. 
We use the spaCy Sentencizer \cite{spacySent} for sentence segmentation.

\subsection{Historical News: ProQuest}

For historical news data, we collect news articles from local, regional, and national news publications from the ProQuest database spanning 1960-1980. See Appendix \ref{sec:proquest-newspapers} for a list of the included publications. We chose this historical period because of the high levels of inflation that occurred throughout it, presenting an interesting opportunity to explore inflation narratives.  The final dataset, including ``inflation'' keyword filtering, contains 392,475 articles and 751,380 sentences. We used the BlingFire \citep{blingFire} sentence segmentation tool, as the spaCy Sentencizer did not work well on this historical data.

\subsection{Human Labeling}

\begin{table}[htbp]
    \centering
    \begin{subtable}[b]{0.45\textwidth}
        \centering
        \begin{tabular}{m{2.5cm}rr}
\toprule
\multicolumn{1}{l}{} & \multicolumn{1}{c}{Historical} & \multicolumn{1}{c}{Contemporary}  \\
\midrule
Train / Test & 999 / 488 & 1,119 / 201  \\
Median Words Per Sentence & 26 & 25 \\ 
\bottomrule
\end{tabular}

        \caption{Human annotation train and test set sizes, and median sentence lengths.}
        \label{tab:annotated-splits}
    \end{subtable}
    \hfill
    \vspace{10pt}
    \begin{subtable}[b]{0.45\textwidth}
         \centering
        \begin{tabular}{lrr}
\toprule
\multicolumn{1}{l}{Dataset} & \multicolumn{1}{c}{Binary} & \multicolumn{1}{c}{Multi-class}  \\
\midrule
Contemporary & 0.67 & 0.59 \\
Historical & 0.80 & 0.66 \\
\bottomrule
\end{tabular}

        \caption{Test set Inter-annotator agreement: Krippendorff's alpha using MASI distance weighting (\citealp{hayes2007answering})}
        \label{tab:agreement}
    \end{subtable}
    \caption{Human annotation statistics}
    \label{tab:maintable}
\end{table}

Three members of our team manually annotated training and test sets. In Table \ref{tab:annotated-splits} we report the sizes of our train and test splits. We targeted train sets of approximately 1,000 examples. This provided us with sufficient training data for model fine-tuning. For the test sets, all three annotators label the same subset of data. For ProQuest, annotators initially labeled a test set of 500 sentences, however, this is reduced to 488 after filtering out texts longer than 150 words when the sentence segmentation failed.

Table \ref{tab:agreement} shows a moderate to high degree of agreement for a pragmatic annotation task, across both the historical and contemporary news annotations. We hypothesize that historical news agreement is higher than contemporary news due to (1) annotators having had more experience with the annotation since the historical annotation came second, and (2) less variation in the sourcing of historical news. The historical ProQuest news dataset primarily contains a collection of professional news publications, which results in less linguistic novelty and variation. In contrast, the contemporary news in the NOW corpus comes from a far greater variety of online sources. This variation could cause a more difficult annotation task. We present an analysis of annotator disagreement in section \ref{sec:error}. See Appendix \ref{sec:interface-examples} for annotation interface examples.

\subsection{Descriptive Statistics}
\label{sec:desc-stats}

We focus on \textit{causal micro-narratives} to ensure that we distinguish between general mentions of inflation in news text and a more targeted framing that presents causal stories about inflation. Analysis of the human annotations reveals that 49\% and 47\% of the contemporary and historical news sentences, respectively, were labeled as non-narratives. Given that these sentences are already keyword-filtered to include \textit{inflation}, this amounts to a significant fraction of them and supports the intent of our definition and annotation scheme.

\begin{figure}[ht]
    \centering
    \begin{subfigure}[b]{0.49\textwidth}
        \centering
        \includegraphics[width=\textwidth]{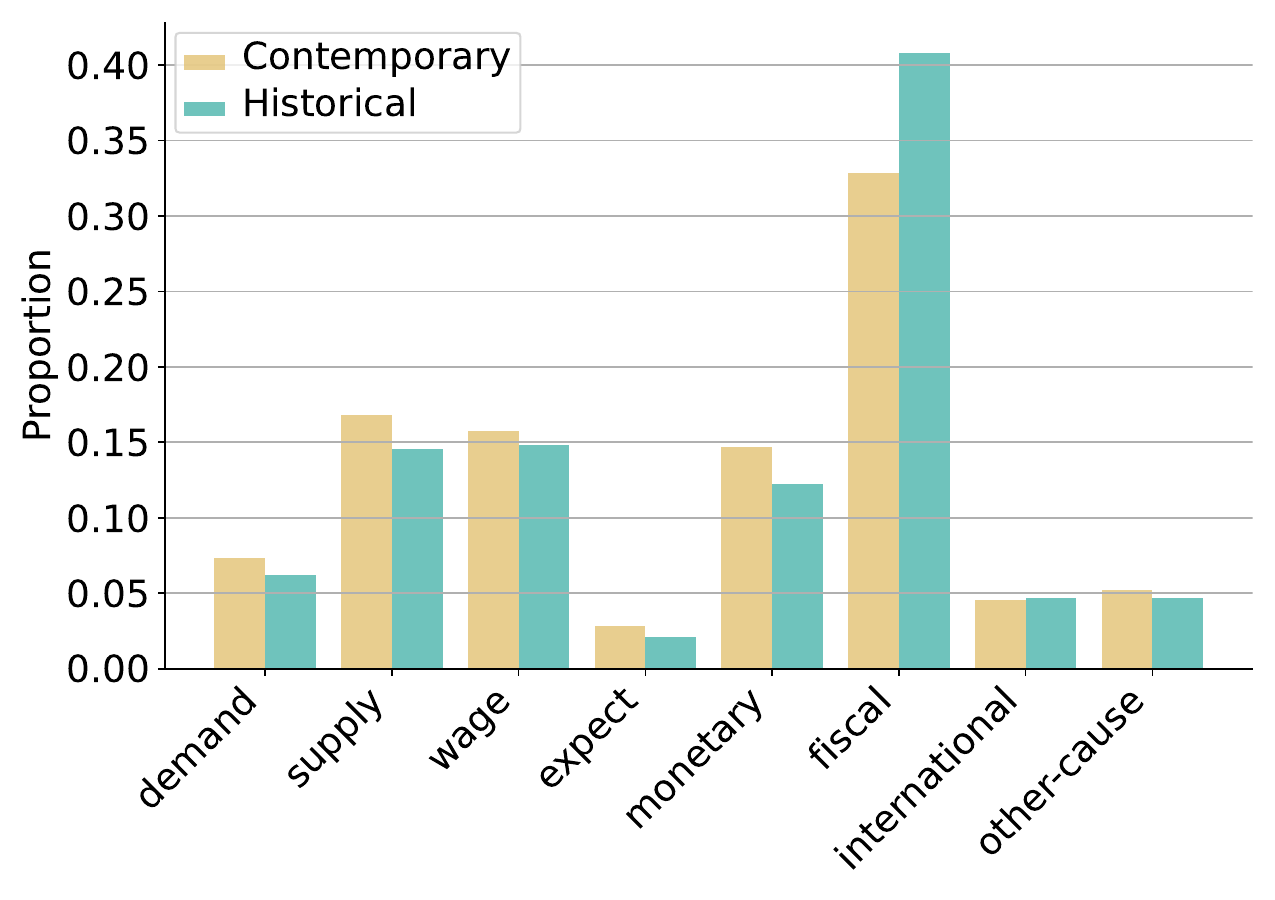}
        \caption{Inflation cause narratives.}
        \label{fig:causes-stats}
    \end{subfigure}
    \hfill
    \begin{subfigure}[b]{0.49\textwidth}
        \centering
        \includegraphics[width=\textwidth]{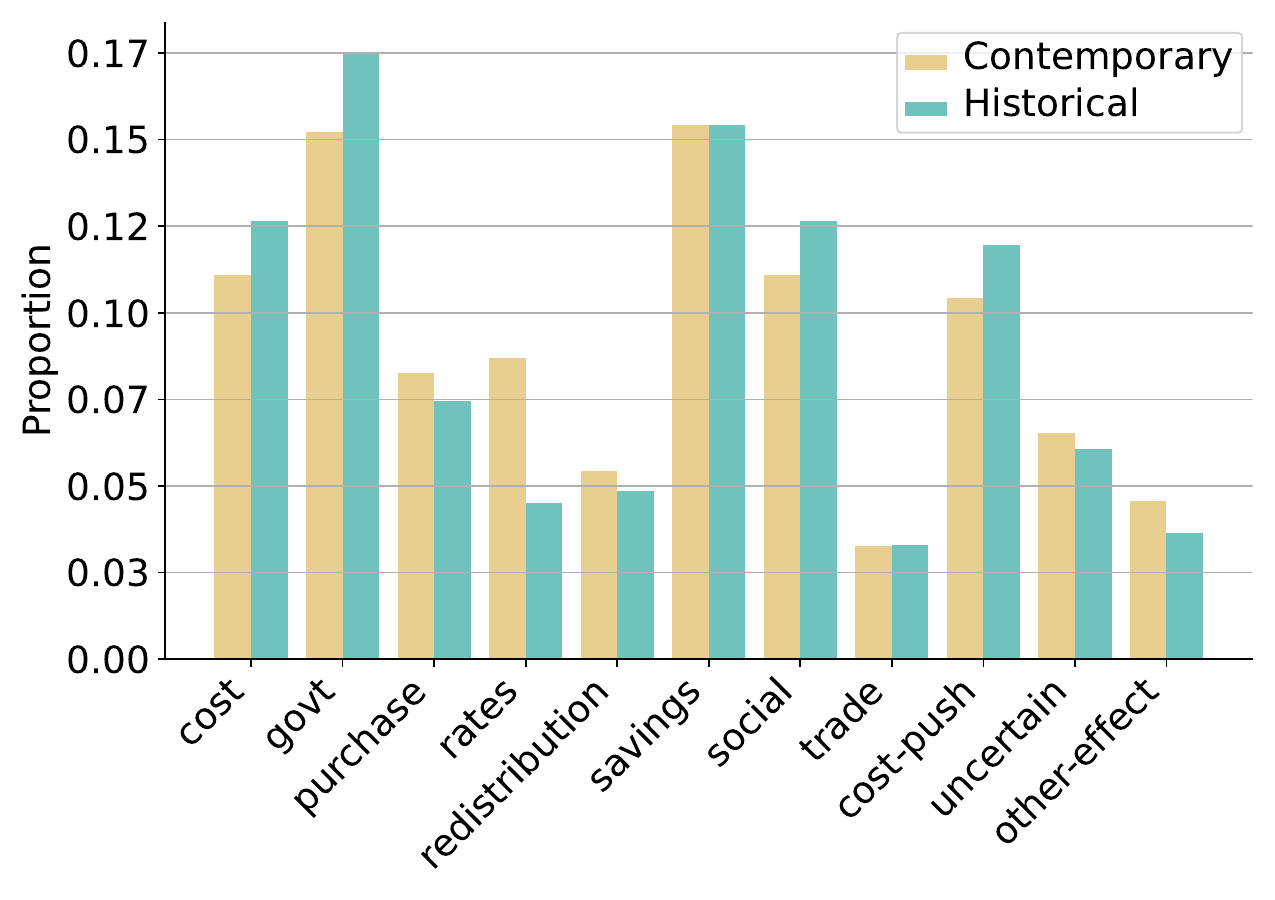}
        \caption{Inflation effect narratives.}
        \label{fig:effects-stats}
    \end{subfigure}
    \caption{Proportions of narrative classes in human annotations. This data combines both the train and test sets. For the test set, majority vote is used to identify one annotation instance.}
    \label{fig:annotations-stats}
\end{figure}

The distribution and prevalence of cause and effect narratives remains largely consistent across human annotations of both datasets. As Figure \ref{fig:annotations-stats} shows, there are only small  variations between most labels. Exceptions include \textit{fiscal} and \textit{govt}, which are more prevalent in historical news, and \textit{rates}, which occurs more frequently in the contemporary data. These outliers reflect overall differences between inflation-related news in the 1960s and 1970s compared to the 2010s. These particular differences can likely be attributed to the fact that interest rate adjustment as a response to inflation did not become a significant tool deployed by the Federal Reserve until Paul Volcker's tenure as Chairman of the Fed in the 1980s \citep{siegel1998stocks}. As such, during the 60s and 70s, government spending and its relationship to inflation (\textit{fiscal}, \textit{govt}) was a more common topic of discussion.

\section{Methods}

To determine the most effective approach to classify narratives, we compare the performance of LLMs on our classification task for both in-context learning and fine-tuning settings. We focus on these two settings  We format the annotations associated with each sentence as JSON to facilitate automatic processing (see Appendix \ref{sec:ft-prompt}). The LLMs are evaluated on their classification output, expected to be in JSON as well.
We conduct separate experiments with the contemporary and historical data and train separate models for each dataset.

\subsection{In-Context Learning}

LLMs have been shown to be effective in-context, or \textit{few-shot}, learners \citep{llmsfewshotlearners}, so we tested GPT-4o in this setting by providing definitions for all the labels along with 24 narrative classification examples, one for each distinct cause and effect, as well as 5 examples of non-narratives. We use greedy decoding and do not constrain the generation in any way, but find that GPT-4o reliably generated JSON in the correct format.

\subsection{Fine-tuning}

The second modeling approach we evaluate is fine-tuning two open-source, pre-trained LLMs: Llama 3.1 8B (\texttt{meta-llama/Meta-Llama-3.1-8B}) and Phi-2 (\texttt{microsoft/phi-2}). We chose these two models because they represent high quality LLMs that have performed well on LLM benchmarks. Additionally, because of their relatively smaller parameter counts compared to other recent LLMs, they are well suited for efficient inference at scale. Indeed, while this classification task test set is relatively small, the ultimate aim of our work is to enable researchers to do complex narrative classification tasks at the scale of millions of sentences from news articles across long time horizons. 

For fine-tuning, the input consists of the possible causes and effects, their definitions, and a brief instruction. We include the full fine-tuning prompt in Appendix \ref{sec:ft-prompt}. We follow standard auto-regressive language modeling but only back propagate the language modeling loss for tokens associated with binary and multi-class labels, rather than other tokens associated with the JSON notation.
We use LoRA-based Parameter-Efficient Fine-Tuning (PEFT)~\citep{hu2021lora} to train a subset of the parameters. See Appendix \ref{sec:hyperparams} for fine-tuning hyper-parameters.

In few- and zero- shot experiments both models achieved extremely low F1 scores (0.12 or lower). As a result, for the purposes of this work, we focus on evaluating fine-tuned versions of the two open-source models, rather than their zero-shot performance. 

\subsection{Evaluation}
We evaluate each aspect of a narrative classification separately using micro-averaged F1 scores. We use micro averaging, rather than weighted- or macro- averaging to get an overall picture of model performance across all instances, including less represented classes. Micro-averaged scores use the standard binary-F1 score formula, but, importantly, the precision and recall scores are based on true and false positives across all instances, irrespective of individual class distinctions. Because each sentence could have narratives with multiple causes and/or effects, micro-averaged F1 differs from a regular accuracy score. 

To resolve disagreements between annotators in the test set, we use majority rule to identify gold-labels. In practice, 97\% of the test set instances have agreement between at least two annotators, allowing us to retain almost the entire test set for evaluation.

\section{Results}

\begin{table}[h]
\centering
\begin{tabular}{lrrr}
\toprule
& \multicolumn{1}{c}{Llama3.1} & \multicolumn{1}{c}{Phi-2} & \multicolumn{1}{c}{GPT-4o}  \\
\midrule
\multicolumn{3}{l}{\textit{Binary}} \\
\hspace{0.5cm}Hist.  & 0.78  & \textbf{0.83} & 0.47  \\
\hspace{0.5cm}Contemp. & \textbf{0.87} & 0.79 & 0.63  \\
\multicolumn{3}{l}{\rule{0pt}{3ex} \textit{Multiclass}} \\
\hspace{0.5cm}Hist. &  \textbf{0.62} & 0.60 & 0.46 \\
\hspace{0.5cm}Contemp. & \textbf{0.71}  & 0.65 &  0.57  \\
\bottomrule
\end{tabular}

\caption{Summary F1 scores for the inflation narrative classification task on Historical (Hist.) and Contemporary (Contemp.) datasets. Phi-2 and Llama 3.1 8B are fine-tuned on a combined dataset totalling 2,118 instances. F1 uses micro-averaging for multi-class and binary for narrative detection. All scores are calculated using majority vote between the three annotators as ground truth. 14 test set instances with no majority annotation are ignored in this score. \textbf{Bolded} values indicate the best performing model on each task (binary and multiclass) and each test set (Historical and Comptemporary).}
\label{tab:model_comp_f1_main}
\end{table}

\begin{table*}[t]
\centering
\begin{tabular}{lccccccccc}
\toprule
& \multicolumn{2}{c}{Llama3.1 8B} & \multicolumn{2}{c}{Phi-2} & \multicolumn{2}{c}{GPT-4o}  \\
\cmidrule(lr){2-3} \cmidrule(lr){4-5} \cmidrule(lr){6-7} \cmidrule(lr){8-9}
\diagbox[width=3cm]{Train}{Test}& Hist. & Contemp. & Hist. & Contemp. & Hist. & Contemp. \\
\midrule
\multicolumn{3}{l}{\textit{Binary}} \\
\hspace{0.5cm}Hist. & 0.64 & 0.75 & 0.75 & 0.82  & 0.47 & 0.70  \\
\hspace{0.5cm}Contemp. & 0.73 & 0.82 & 0.75 & 0.83 & 0.51 & 0.63  \\
\hspace{0.5cm}Hist. + Contemp. & 0.78 & \textbf{0.87} & \textbf{0.83 }& 0.79 & 0.39 & 0.43  \\
   
\multicolumn{3}{l}{\rule{0pt}{3ex} \textit{Multiclass}} \\
\hspace{0.5cm}Hist. & 0.55 & 0.59 & 0.57 & 0.63 & 0.46 & 0.60  \\
\hspace{0.5cm}Contemp. & 0.52 & 0.63 & 0.53 & 0.66 & 0.48 & 0.57  \\
\hspace{0.5cm}Hist. + Contemp. & \textbf{0.62} & \textbf{0.71} & 0.60 & 0.65 & 0.43 & 0.46  \\
\bottomrule
\end{tabular}

\caption{F1 scores for the inflation narrative classification task on Historical (Hist.) and Contemporary (Contemp.) Datasets. Phi-2 and Llama 3.1 8B are fine-tuned. F1 uses micro-averaging for multi-class and binary for narrative detection. All scores are calculated using majority vote between the three annotators as ground truth. 14 test set instances with no majority annotation are ignored in this score. Columns specify the datasets used for training; and rows, the results on test sets. \textbf{Bolded} values indicate the best performing model and training data combination for each task (binary and multiclass) and each test set (Historical and Comptemporary).}
\label{tab:model_comp_f1}
\end{table*}

We compare model performance in Table \ref{tab:model_comp_f1_main}. Fine-tuned Llama 3.1 8B performs the best and, along with Phi-2, outperforms GPT-4o. GPT-4o particularly suffers on Historical data and the binary narrative detection overall.

To better understand how models trained on these datasets may generalize to news from other periods, we present in Table \ref{tab:model_comp_f1} a breakdown of model performance in several training and evaluation settings. First, we evaluate how well models fine-tuned on Historical and Contemporary data perform on corresponding held-out data, assessing in-domain generalization. Second, we compare how well models generalize to out-of-distribution (OOD) data by evaluating performance on Historical data when trained on Contemporary data, and vice-versa. Finally, we combine both the historical and contemporary data during the learning phase and evaluate performance on the individual datasets, revealing how well models can learn from the additional data despite the domain-shift.

\subsection{In-Domain Generalization}

When trained and evaluated on the same individual dataset, Phi-2 outperforms other models. Interestingly, however, Llama 3.1 8B is better able to learn from both the Historical and Contemporary datasets, exhibiting impressive improvements of up to 14\%, despite the 50-year gap between the news in the two datasets. In contrast, Phi-2 struggles and even degrades in performance on Contemporary data multi-class classification. All models perform better on contemporary data, likely because recent text and language from 2012-2023 are more prevalent in their pre-training corpora than historical newspaper data.

\subsection{Out-of-Domain Generalization}

On the multiclass narrative classification task, a common pattern emerges across both fine-tuned models. We observe that test set performance degrades by 3-4\% on OOD data relative to in-domain data. This represents a moderate drop in performance and could be attributed to changes in the distribution of narratives across the Historical and Contemporary datasets, as explained in Section \ref{sec:desc-stats} and Figure \ref{fig:annotations-stats}. In contrast, the binary prediction task reveals a different effect. Phi-2 performs the same regardless of which dataset is used for training and which is used for testing but Llama 3.1 8B achieves up to an 11\% improvement on narrative detection in Historical news sentences when trained on the Contemporary data. In the reversed setting, Llama 3.1 8B performance degrades by 7\%. This pattern suggests that training Llama on Contemporary data is more successful than Historical data.

\subsection{Error Analysis}

\begin{table*}[t]
\centering
\small
\definecolor{forestgreen}{RGB}{34,139,34}

\begin{tabular}{p{10cm}cc}
\toprule
Sentence & Llama 3.1 8b & Majority Annotation \\
\midrule
"The corrosive effects of inflation eat away at the ties that bind us together as a people," said President Carter Thursday in the third of the messages--the budget, the State of the Union, and the Economic Report--that make up the traditional January triad. & \textcolor{red}{no-narrative} & \textcolor{forestgreen}{social}  \\
\midrule
But he acknowledged that the Administration-projected rate of 6.5\% to 7\% inflation this year still made it the nation s worst domestic problem. & \textcolor{red}{no-narrative} & \textcolor{forestgreen}{social} \\
\midrule
He said inflation was every American's problem and that the nation's economic, military and spiritual strength depended on solving it. & \textcolor{red}{no-narrative} & \textcolor{forestgreen}{social} \\
\midrule
'They have and will cause Inflation to accelerate in the state and the Chicago area, destroy jobs that otherwise would be available, lower family income, and increase taxes,"he said. & \textcolor{red}{fiscal} & \textcolor{forestgreen}{govt, purchase, cost-push} \\
\midrule
"Inflation has slowed, but people's perception of that changes," he said. & \textcolor{orange}{no-narrative} & \textcolor{orange}{expect} \\
\midrule
Carter finally became convinced that inflation was the No. 1 problem. & \textcolor{orange}{no-narrative} & \textcolor{orange}{govt} \\
\midrule
Consequently, increases in valuation due to inflation do indeed raise the number of actual dollars in property taxes owed. & \textcolor{orange}{govt} & \textcolor{orange}{savings} \\

\bottomrule
\end{tabular}

\caption{Comparison of fine-tuned LLama 3.1 8B and human annotations.}
\label{tab:nonarrative-error}
\end{table*}

To better understand model performance on this task and the variation between fine-tuning a smaller LLM and few-shot prompting a large propriertary LLM, we conduct a fine-grain analysis of the individual narrative classification predictions as well as an analysis of the three sets of human annotations to better understand the disagreements that exist between them and how those disagreements may related to model prediction errors. As the best performing LLM overall, we focus on Llama 3.1 8B (henceforth, \textit{Lllama}) and compare it to GPT 4o, the only propriertary model in our experiments.

\paragraph{Human Annotator Disagreements}
By majority rule, our three human annotators find partial agreement on 474 out of 488 test set instances, and full agreement on 471. While this is a higher rate of majority agreement, there are nonetheless non-negligible disagreements between individual annotators. Since we use training data sourced from each annotator individually, understanding these disagreements can contextualize how model performance is impacted. Most annotator disagreements stem from differing judgments on narrative presence, not category assignment. Annotators rarely clash over which specific narrative category to apply, but often diverge on whether a narrative exists in the text at all. Furthermore, certain annotators are systematically more likely to detect narratives than others, driving this specific form of disagreement. 

\paragraph{Hallucinating Narratives}
Fine-tuning is effective at teaching a model to distinguish between narratives and non-narratives, compared to in-context learning. GPT-4o, which was not fine-tuned, correctly classifies roughly 47\% and 60\% fewer non-narratives in the contemporary NOW and historical ProQuest test sets, respectively, than Llama. Despite extensive experimentation with different prompts, we consistently observed that GPT-4o struggled to understand the distinction we stipulate between narratives and non-narratives. We can likely attribute this to our precise definition of narrative, such that these otherwise highly capable LLMs have limited in-context demonstration data to draw on to learn this capability.

\paragraph{Natural Variation \& Ambiguity in Language}
Table \ref{tab:nonarrative-error} presents several instances where Llama predictions did not match the human labels. 
The first three examples illustrate that Llama's impressive 0.87 F1 score on binary narrative detection comes at the cost of false negative predictions. In fact, these three instances of failing to predict \textit{Social \& Political Impact (social)} are representative of the most common type of false negative error in Llama predictions. Interestingly, annotating \textit{social} or not is the most common disagreement of this type among the annotators. Nonetheless, the three examples in Table \ref{tab:nonarrative-error} show failures of Llama to identify the implied, yet clear, references to inflation's social and political impact.

In contrast, the final four examples demonstrate the natural ambiguity and difficulty inherit in this task. Consider the fourth sentence. While to a human, it may be quite natural to understand this sentence as inflation being the cause of the job destruction, lower family income, and increased taxes, it is not explicit in the sentence. In fact, the more explicit mention of causation in the sentence is ``they have and will cause inflation". Llama predicts a \textit{cause of inflation} narrative (``fiscal"), whereas the reference labels are \textit{effects of inflation} (``govt, purchase, cost-push"). In practice, this sentence does not mention who ``they" is referring to, so the prediction, while a reasonable guess, is not supported. The final three examples show scenarios where the Llama predictions and human annotations could both be considered correct, depending on one's perspective. All these examples illustrate the challenging nature of the task and the natural variation that is inherent to it.

\section{Conclusion}

This paper proposes a \textit{causal micro-narrative} classification task. By developing a comprehensive classification scheme and leveraging both fine-tuned and few-shot prompted large language models, we demonstrate the feasibility of automating the detection and categorization of these narratives at scale. Our results show that fine-tuned models, particularly Llama 3.1 8B, outperform few-shot prompted models in distinguishing between narrative and non-narrative content, while maintaining competitive performance in classifying specific narrative types.

The error analysis reveals that the task of identifying causal micro-narratives is inherently complex, with natural ambiguities in language and variation in human interpretations. Despite these challenges, our approach provides a foundation for future research in narrative analysis within the social sciences. By enabling the systematic extraction of causal narratives from large-scale textual data, this work opens up new possibilities for studying the evolution and impact of narratives over time, potentially offering valuable insights for policymakers, economists, and social scientists alike.

\section{Limitations}
The method we propose for extracting and classifying \textit{causal micro-narratives} requires the manual development of an ontology of causes and effects for any new target. This limits automated data-driven discovery of new narratives (i.e., causes and effects not already pre-established). However, the binary micro-narrative detection task included in this paper may be helpful in filtering a large corpus into a smaller dataset of sentences that contain narratives. This may facilitate discovering new narratives, either manually, or with an automated method. In this paper, we do not evaluate this use-case but we believe this to be a good direction for future work.

\bibliography{anthology,custom}

\clearpage
\onecolumn
\appendix
\section*{Appendix}
\section{ProQuest Newspapers}
\label{sec:proquest-newspapers}

Chicago Tribute, Chicago Defender, Los Angeles Times, Los Angeles Sentinel, Atlanta Daily World, Cleveland Call and Post, Detroit Free Press, Indianapolis Star, Kansas City Call, Louisville Courier Journal, Louisville Defender, Michigan Chronicle, Minneapolis Star Tribune, New York Amsterdam News, New York Tribute / Herald Tribune, Norfolk Journal and Guide, Philadelphia Tribune, Pittsburgh Courier, Pittsburgh Post-Gazette, San Francisco Chronicle, St. Louis American, St. Louis Post Dispatch, The Baltimore Afro-American, The Boston Globe, The Christian Science Monitor, The Cincinnati Enquirer, The Nashville Tennessean, The New York Times, The Wall Street Journal, The Washington Post, U.S. Newsstream, U.S. Major Dailies.

\clearpage
\section{Classification Task}
\label{sec:cls-task}
\begin{table}[ht!]
\centering
\resizebox{\columnwidth}{!}{
\begin{tabular}{p{3.5cm}cp{8cm}}
\toprule
\textbf{Narrative} & \textbf{Label} & \textbf{Definition Excerpt} \\
\midrule
\multicolumn{3}{l}{\textit{Causes}} \\
\midrule
Demand-side Factors & demand & Pull-side or demand-pull inflation. \\
Supply-side Factors & supply & Push-side or cost-push inflation. \\
Built-in Wage Inflation & wage & Also known as wage inflation or wage-price spiral.  \\
Monetary Factors & monetary & Central bank policies that contribute to inflation. \\
Fiscal Factors & fiscal & Government policies that contribute to inflation. \\
Expectations & expect & The expectation that inflation will rise often leads to a rise in inflation. \\
International Trade \& Exchange Rates & international & International trade and exchange rate factors that can cause inflation. \\
Other Causes & other-cause & Causes not included in above. \\
\midrule
\multicolumn{3}{l}{\textit{Effects}} \\
\midrule
Reduced Purchasing Power & purchase & Inflation erodes the purchasing power of money (such as the U.S. dollar) over time. \\
Cost of Living Increases & cost & Inflation can raise the cost of living, particularly impacting individuals on fixed incomes, pensioners, and those with lower wages. \\
Uncertainty Increases & uncertain & Inflation can create uncertainty about future prices (or future inflation itself), particularly if the inflation is high or unpredictable. \\
Interest Rates Raises & rates & Central banks may respond to inflation by raising interest rates to curb spending and investment. \\
Income or Wealth Redistribution & redistribution & Inflation can redistribute income and wealth between people in the economy. \\
Impact on Savings & savings & Inflation can affect various types of savings/financial investments. \\
Impact on Global Trade & trade & Inflation can impact a country's trade or competitiveness in global markets. \\
Cost-Push on Businesses & cost-push & Rising costs of production due to inflationary pressures can squeeze business profits, potentially leading to reduced investment, job cuts and unemployment, or higher prices for consumers. \\
Social and Political Impact & social & Inflation can have social and political economic implications. \\
Government Policy \& Public Finances Impact & govt & Inflation may impact government spending policies or programs. \\
Other Effects & other-effect & Effects not included in above. \\
\bottomrule
\end{tabular}

}
\caption{Narrative categories, their label used in the classification task, and an excerpt of their definitions. These categories were selected and define by a domain expert, using a combination of domain knowledge, google searches, and LLM interactions. When using a LLM (Open AI ChatGPT 3.5, Google Bard/Gemini, Anthropic Claude), the prompt was “what are the causes (effects) of inflation? Describe the economic mechanisms and give examples”. If we wanted to expand on a cause (effect), the prompt was “explain economic mechanisms and examples of xxxx as a cause (effect) of inflation”. We also relied on Google searches of “causes (effects) of inflation”.}

\label{tab:narratives}
\end{table}

\clearpage

\section{Annotation Interface}
\label{sec:interface-examples}
\begin{figure*}[h]
    \centering
    \includegraphics[width=1\textwidth]{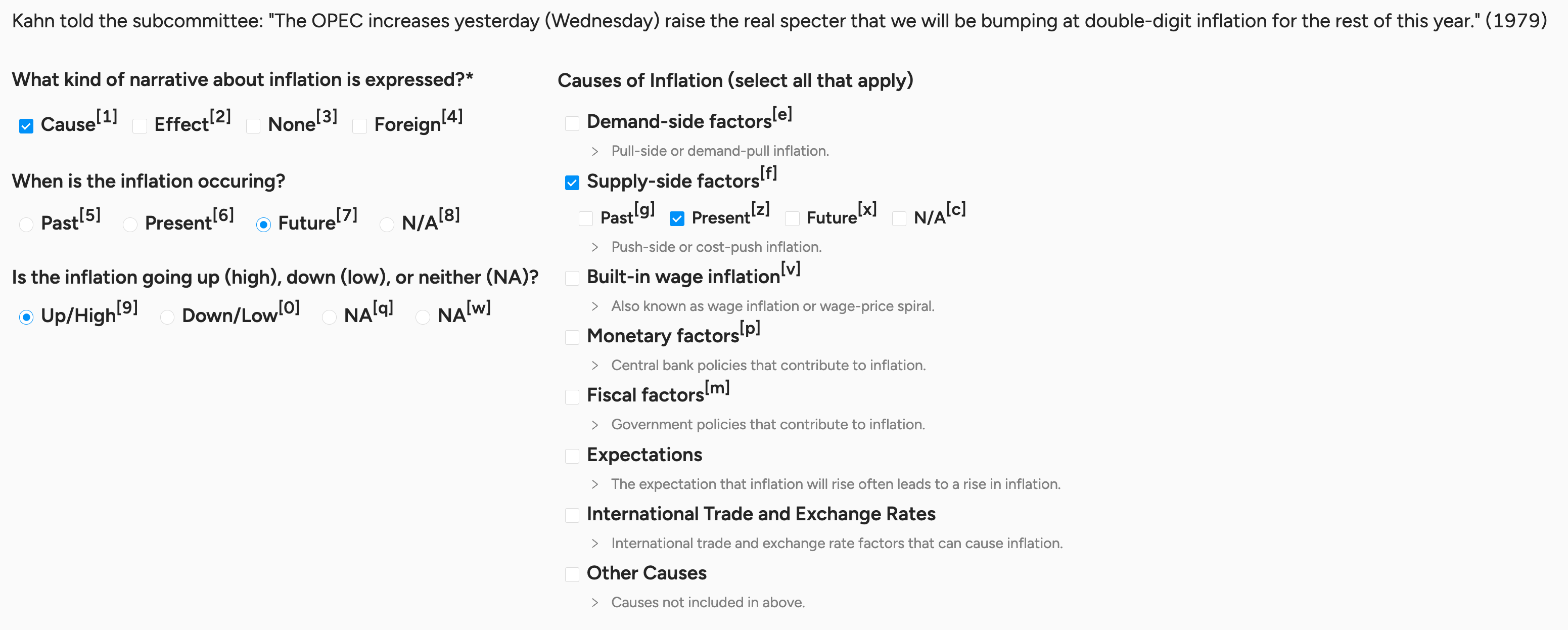}
    \caption{Example of an annotation for a narrative about the cause of inflation.}
    \label{fig:ex_cause}
\end{figure*}

\begin{figure*}[h]
    \centering
    \includegraphics[width=1\textwidth]{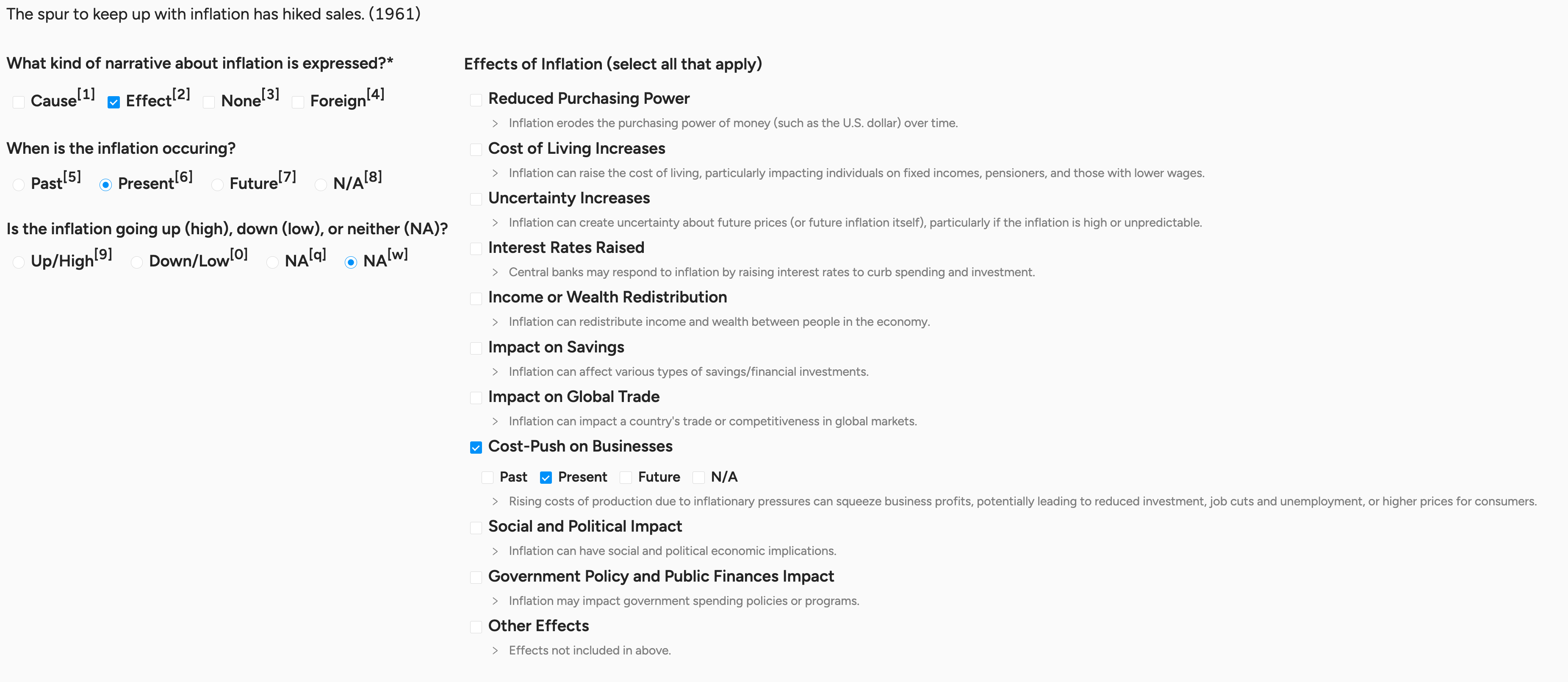}
    \caption{Example of an annotation for a narrative about the effect of inflation.}
    \label{fig:ex_effect}
\end{figure*}

\section{LLM Prompts and Inputs}
\label{sec:ft-prompt}

\lstset{language=Java, basicstyle=\ttfamily\footnotesize}
\begin{figure}[ht!]
\begin{lstlisting}[style=pseudocode]
Below are lists of causes and effects of inflation.

Causes of inflation:
[demand] Demand-side factors: Pull-side or demand-pull inflation.
[supply] Supply-side factors: Push-side or cost-push inflation.
[wage] Built-in wage inflation: Also known as wage inflation or wage-price spiral. 
[monetary] Monetary factors: Central bank policies that contribute to inflation.
[fiscal] Fiscal factors: Government policies that contribute to inflation.
[expect] Expectations: The expectation that inflation will rise often leads to a rise in inflation.
[international] International Trade and Exchange Rates: International trade and exchange rate factors that can cause inflation.
[other-cause] Other Causes: Causes not included in above.

Effects of inflation:
[purchase] Reduced Purchasing Power: Inflation erodes the purchasing power of money (such as the U.S. dollar) over time.
[cost] Cost of Living Increases: Inflation can raise the cost of living, particularly impacting individuals on fixed incomes, pensioners, and those with lower wages.
[uncertain] Uncertainty Increases: Inflation can create uncertainty about future prices (or future inflation itself), particularly if the inflation is high or unpredictable.
[rates] Interest Rates Raised: Central banks may respond to inflation by raising interest rates to curb spending and investment.
[redistribution] Income or Wealth Redistribution: Inflation can redistribute income and wealth between people in the economy.
[savings] Impact on Savings: Inflation can affect various types of savings/financial investments.
[trade] Impact on Global Trade: Inflation can impact a country's trade or competitiveness in global markets.
[cost-push] Cost-Push on Businesses: Rising costs of production due to inflationary pressures can squeeze business profits, potentially leading to reduced investment, job cuts and unemployment, or higher prices for consumers.
[social] Social and Political Impact: Inflation can have social and political economic implications.
[govt] Government Policy and Public Finances Impact: Inflation may impact government spending policies or programs.
[other-effect] Other Effects: Effects not included in above.

Identify all causes and effects of inflation that are expressed in the sentence:
\end{lstlisting}
\caption{Causal Micro-Narrative classification prompt. For few-shot/zero-shot prompts, examples are listed before the final sentence.}
\end{figure}

Due to the hierarchical multi-label classification task, we represent a complete narrative classification as JSON. This paper focuses only on the prediction results; i.e., the values associated with the fields ``contains-narrative" and ``narratives". However, our task includes additional information which we will discuss in future work. We define the JSON schema as follows:

\begin{lstlisting}[language=json]
{
  "foreign": true|false,
  "contains-narrative": true|false,
  "inflation-narratives": [
    "inflation-time": "past"|"present"|"future"| "na",
    "inflation-direction": "down"|"up"|"na",
    "narratives": [
        {"causes"|"effect": category, "time": "past"|"present"|"future"| "na"},
        ...
    ]
  ] | null
}
\end{lstlisting}

\section{Hyperparameters}
\label{sec:hyperparams}
\begin{table}[ht!]
        \tiny
        \let\center\empty
        \let\endcenter\relax
        \centering
        \resizebox{1\textwidth}{!}{\begin{threeparttable}
\begin{tabular}{ccccc}
\toprule
\multicolumn{1}{c}{Max Steps} & \multicolumn{1}{c}{Effective Batch Size} & \multicolumn{1}{c}{Optimizer} & \multicolumn{1}{c}{Learning Rate} & \multicolumn{1}{c}{LoRA $r,\alpha$} \\
\midrule
 600 & 16  & AdamW  & 1e-4  & 16, 32 \\
\bottomrule
\\
\end{tabular}
\end{threeparttable}

}
    \caption{Fine-tuning hyper-parameters for Phi-2 and Llama 3.1 8B.}
    \label{tab:lr_hyperparams_LED}
\end{table}

\section{Confusion Matrices}
\label{sec:confusion-matrices}

\label{sec:error}

\begin{figure}[ht!]
    \centering
        \includegraphics[width=\textwidth]{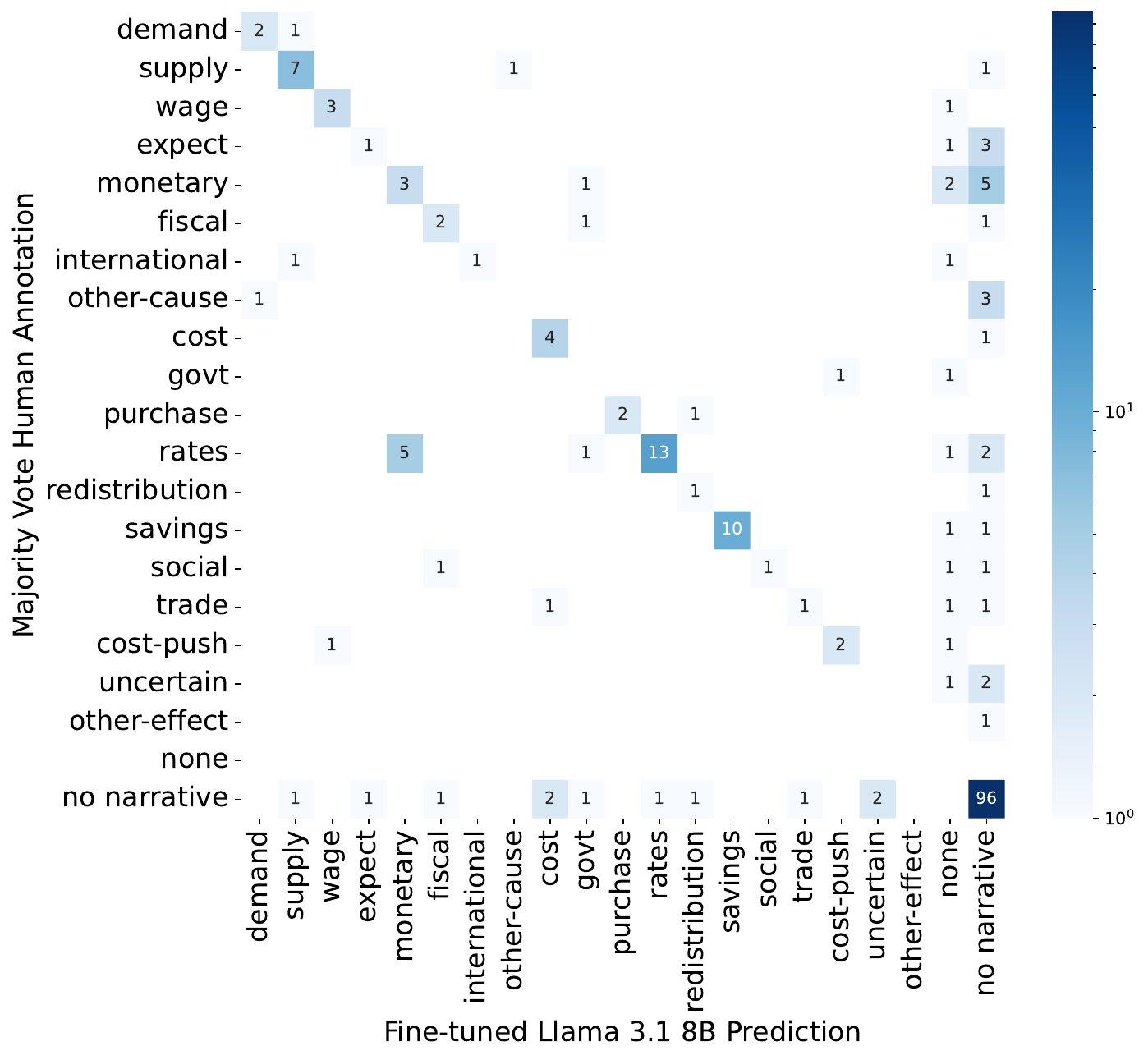}
        \caption{Confusion matrix: NOW Test set fine-tuned Llama 3.1 8B predictions against majority vote human ground-truths. Label ``none" indicates when a narrative does not match any of the narratives in the comparison set. For example, if a model prediction is that a sentence contains a narrative about ``rates" and one about ``monetary" and the human label is ``rates", then ``monetary" would be matched with ``none".}
        \label{fig:phi2_heatmap}
\end{figure}
\begin{figure}[ht!]
        \centering
        \includegraphics[width=\textwidth]{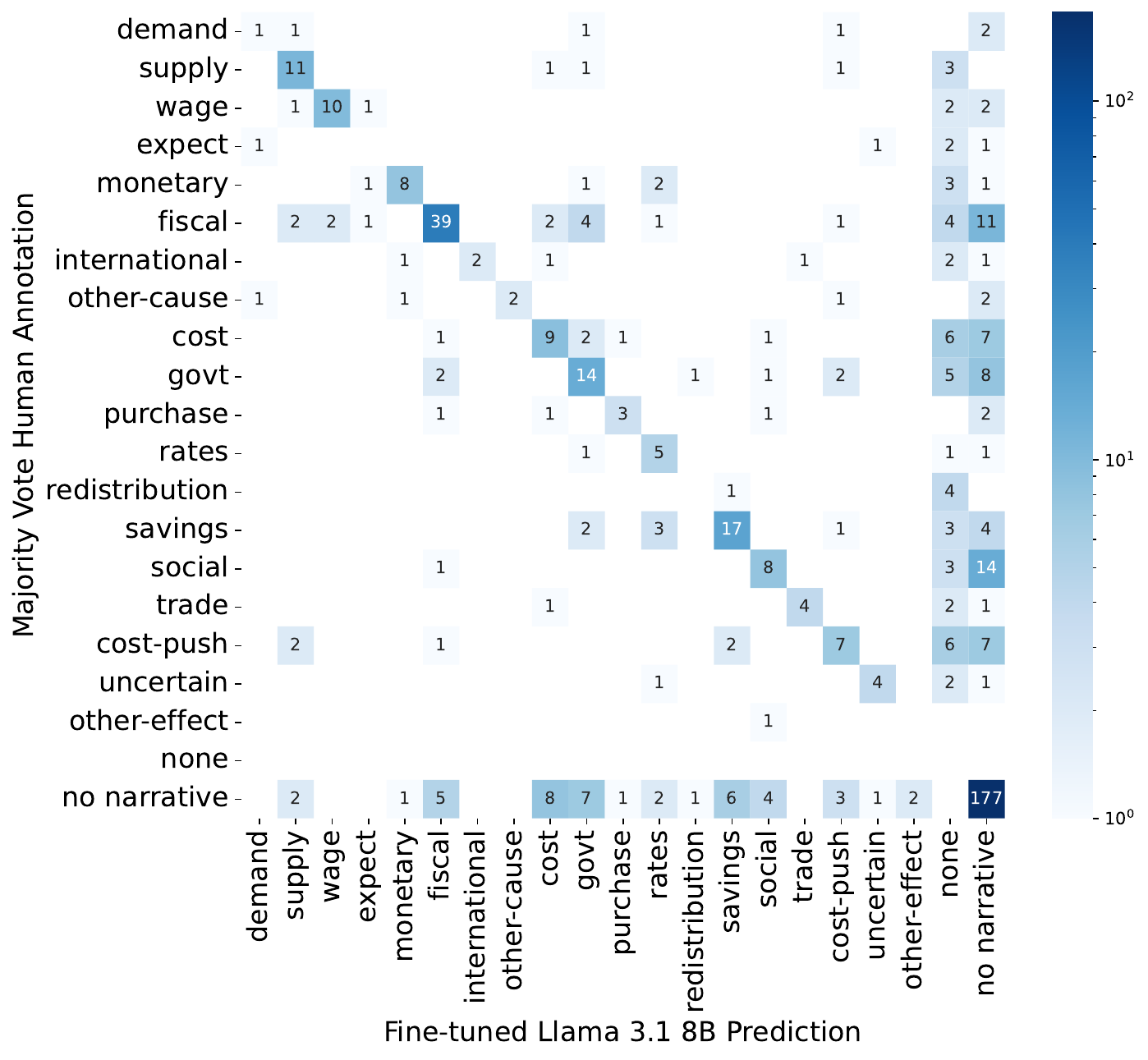}
        \caption{Confusion matrix: ProQuest Test set fine-tuned Llama 3.1 8B predictions against majority vote human ground-truths. Label ``none" indicates when a narrative does not match any of the narratives in the comparison set. For example, if a model prediction is that a sentence contains a narrative about ``rates" and one about ``monetary" and the human label is ``rates", then ``monetary" would be matched with ``none".}
        \label{fig:claude_heatmap_proquest}
    \label{fig:comparison_heatmaps_llama}
\end{figure}

\begin{figure}[ht!]
    \centering
        \includegraphics[width=\textwidth]{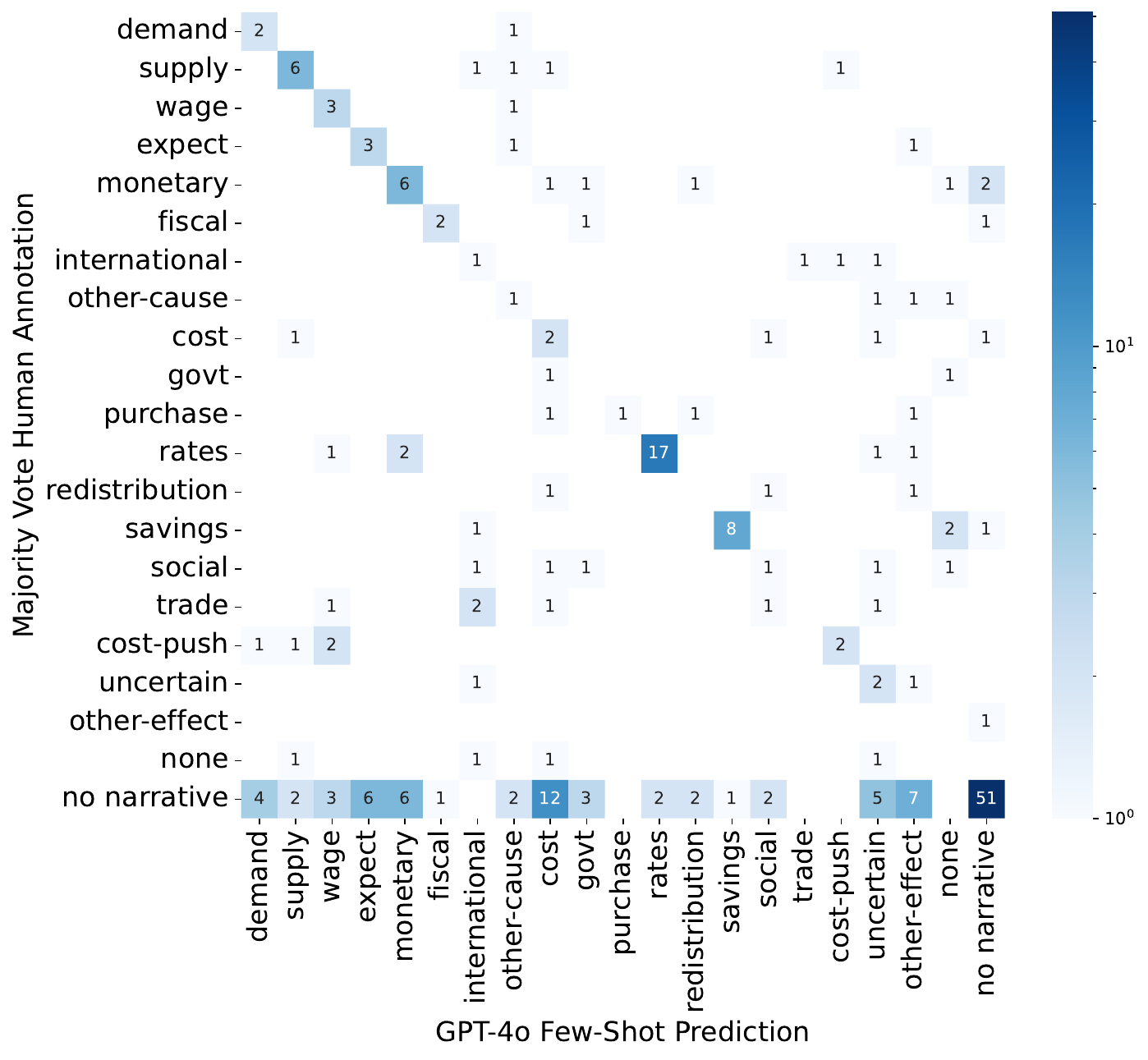}
        \caption{Confusion matrix: NOW Test set GPT-4o predictions against majority vote human ground-truths. Label ``none" indicates when a narrative does not match any of the narratives in the comparison set. For example, if a model prediction is that a sentence contains a narrative about ``rates" and one about ``monetary" and the human label is ``rates", then ``monetary" would be matched with ``none".}
        \label{fig:phi2_heatmap_now}
\end{figure}
\begin{figure}[ht!]
        \centering
        \includegraphics[width=\textwidth]{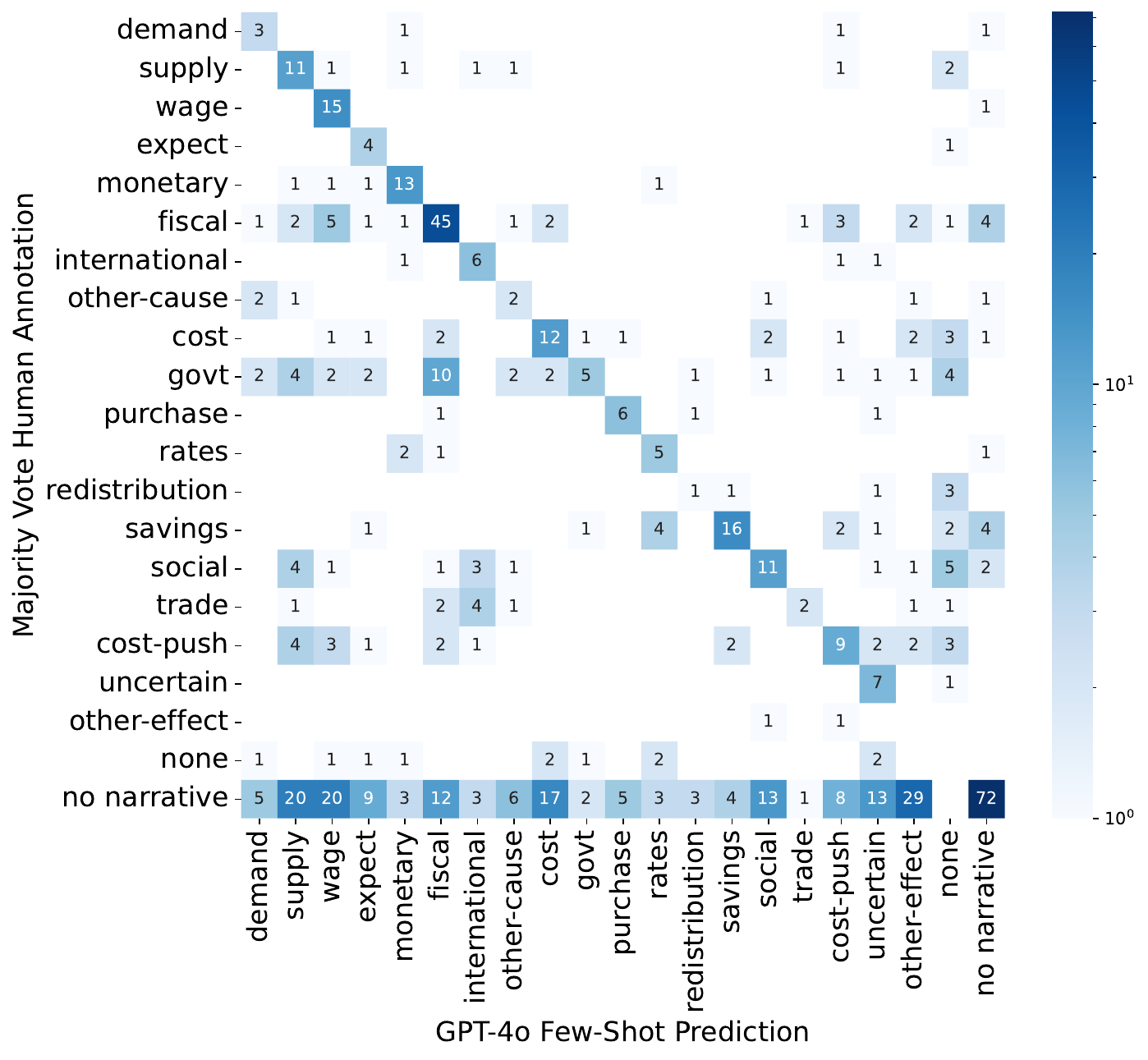}
        \caption{Confusion matrix: ProQuest Test set GPT-4o predictions against majority vote human ground-truths. Label ``none" indicates when a narrative does not match any of the narratives in the comparison set. For example, if a model prediction is that a sentence contains a narrative about ``rates" and one about ``monetary" and the human label is ``rates", then ``monetary" would be matched with ``none".}
        \label{fig:claude_heatmap}
    \label{fig:comparison_heatmaps_gpt4}
\end{figure}

\end{document}